\pdfoutput=1

\documentclass[11pt]{article}

\usepackage{EMNLP2022}

\usepackage{times}
\usepackage{latexsym}

\usepackage[T1]{fontenc}

\usepackage[utf8]{inputenc}

\usepackage{microtype}

\usepackage{inconsolata}
\usepackage{microtype}
\usepackage{times}
\usepackage{soul}
\usepackage{url}
\usepackage{caption}
\usepackage{graphicx}
\usepackage{amsmath}
\usepackage{amsthm}
\usepackage{booktabs}
\usepackage{algorithm}
\usepackage{algorithmic}
\usepackage{multirow}
\usepackage{makecell}
\usepackage{bm}
\usepackage{setspace}
\usepackage{xcolor}
\usepackage{enumitem}

\newcommand{\dqyang}[1]{{\color{black} #1}}
\newcommand{\ysy}[1]{{\color{black} #1}}
\newcommand{\eat}[1]{}

\title{Generative Entity Typing with Curriculum Learning}

\author{Siyu Yuan\textsuperscript{\rm $\heartsuit$}$^1$,
Deqing Yang\textsuperscript{\rm $\heartsuit$}$^2$\thanks{~~Corresponding authors.},
Jiaqing Liang\textsuperscript{\rm $\heartsuit$}$^2$,
Zhixu Li\textsuperscript{\rm $\spadesuit$}$^2$
Jinxi Liu\textsuperscript{\rm $\heartsuit$}$^1$, \\
\bf Jingyue Huang\textsuperscript{\rm $\heartsuit$}$^2$,
Yanghua Xiao\textsuperscript{\rm $\spadesuit\clubsuit$}$^2$\footnotemark[1]\\
\textsuperscript{\rm $\heartsuit$}School of Data Science, Fudan University, Shanghai, China\\
\textsuperscript{\rm $\spadesuit$}Shanghai Key Laboratory of Data Science, School of Computer Science, Fudan University\\
\textsuperscript{\rm $\clubsuit$}Fudan-Aishu Cognitive Intelligence Joint Research Center\\
$^1$\texttt{\{syyuan21,jxliu22\}@m.fudan.edu.cn}, \\
$^2$\texttt{\{yangdeqing,liangjiaqing,zhixuli,jingyuehuang18,shawyh\}@fudan.edu.cn}
}

\begin{document}
\maketitle
\begin{abstract}
Entity typing aims to assign types to the entity mentions in given texts. The traditional classification-based entity typing paradigm has two unignorable drawbacks: 1) it fails to assign an entity to the types beyond the predefined type set, and 2) it can hardly handle few-shot and zero-shot situations where many long-tail types only have few or even no training instances. 
To overcome these drawbacks, we propose a novel generative entity typing (GET) paradigm: given a text with an entity mention, the multiple types for the role that the entity plays in the text are generated with a pre-trained language model (PLM).
However, PLMs tend to generate coarse-grained types after fine-tuning upon the entity typing dataset. In addition, only the heterogeneous training data consisting of a small portion of human-annotated data and a large portion of auto-generated but low-quality data are provided for model training. To tackle these problems, we employ curriculum learning (CL) to train our GET model on heterogeneous data, where the curriculum could be self-adjusted with the self-paced learning according to its comprehension of the type granularity and data heterogeneity. Our extensive experiments upon the datasets of different languages and downstream tasks justify the superiority of our GET model over the state-of-the-art entity typing models. The code has been released on \url{https://github.com/siyuyuan/GET}.
\end{abstract}

\section{Introduction}\label{sec:Introduction}
Entity typing aims to assign types to mentions of entities from a predefined type set,
which enables machines to better understand natural languages and benefit many downstream tasks, such as entity linking~\cite{yang2019learning} and text classification~\cite{chen2019deep}. 
Traditional entity typing approaches follow the classification paradigm to classify (assign) the entity into a predefined set of types, which have the following two unignorable drawbacks.
1) {\em Closed Type Set}:
The classification-based approaches fail to assign the entity to the types out of the predefined set.
2) {\em Few-shot Dilemma for Long-tail Types}: 
Although fine-grained entity typing (FET) and ultra-fine entity typing approaches can classify entities into fine-grained types, they can hardly handle few-shot and zero-shot issues. In fact, there are many long-tail types only having few or even no training instances in the datasets. 
For example, more than 80\% types have less than 5 instances and 25\% types even never appear in the training data from the ultra-fine dataset~\cite{choi2018ultra}.

\begin{figure}[t]
    \centering
    \includegraphics[width=\columnwidth]{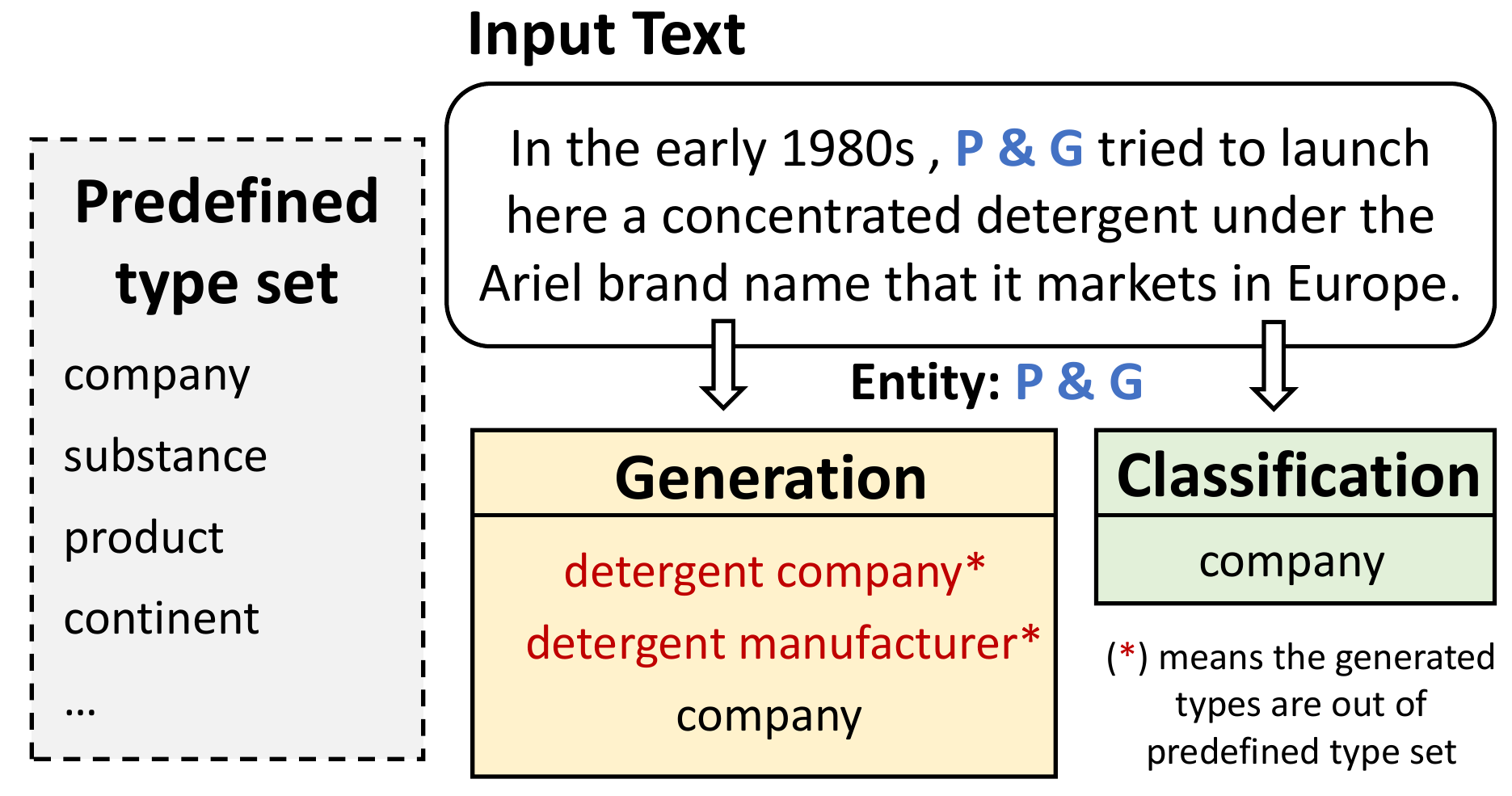}
    \caption{A toy example of entity typing through generation and classification paradigm, respectively.}
    \label{fig:Example}
\end{figure}

To address these drawbacks, in this paper, we propose a novel \textit{generative entity typing} (GET) paradigm: given a text with an entity mention, the multiple types for the role that the entity plays in the text are generated by a pre-trained language model (PLM). 
Compared to traditional classification-based entity typing methods, PLM-based GET has two advantages. First, instead of a predefined closed type set, PLMs can generate more open types for entity mentions due to their strong generation capabilities. 
For example, in Figure~\ref{fig:Example}, fine-grained types such as ``large detergent company" and ``large detergent manufacturer" can be generated by PLMs for entity \textit{P\&G}, which contain richer semantics but are seldom included by a predefined type set. Second, PLMs are capable of conceptual reasoning and handling the few-shot and zero-shot dilemma~\cite{Hwang2021COMETATOMIC2O}, since massive knowledge has been learned during their pre-training.

However, it is nontrivial to realize PLM-based GET due to the following challenges:
1) Entity typing usually requires generating fine-grained types with more semantics, which are more beneficial to downstream tasks. 
However, PLMs are biased to generate high-frequency vocabulary in the corpus due to their primary learning principle based on statistical associations.
As a result, a typical PLM tends to generate high-frequent but coarse-grained types even if we carefully finetune the PLM on the fine-grained entity typing dataset (refer to Figure~\ref{fig:Distribution} in Section~\ref{sec:Experiment}). 
Therefore, how to guide a PLM to generate high-quality and fine-grained types for entities is crucial. 
2) It is costly for humans to annotate a great number of samples with fine-grained types. Therefore, most existing works adopt heterogeneous data consisting of a small portion (less than 10\%) of human-annotated data and a large portion (more than 90\%) of auto-generated low-quality data (e.g., by distant supervision), which greatly hurts the performance of entity typing models~\cite{gong2021abusive}.
How to train a PLM to generate desirable types on these low-quality heterogeneous data is also challenging.

\ysy{
The difficulty of using PLMs to generate high-quality fine-grained types based on the low-quality heterogeneous training data motivates us to leverage the idea from curriculum learning (CL)~\cite{bengio2009curriculum}, which better learns heterogeneous data by ordering the training samples based on their quality and difficulty~\cite{kumar2019reinforcement}. 
In this paper, we propose a CL-based strategy to train our GET model.
Specifically, we first define a fixed {\em curriculum instruction} and partition the training data into several subsets according to the granularity and heterogeneity of samples for model training. 
Based on the curriculum instruction, CL can control the order of using these training subsets from coarse-grained and lower-quality ones to fine-grained and higher-quality ones. 
However, a fixed curriculum ignores the feedback from the training process. 
Thus, we combine the predetermined curriculum with self-paced learning (SPL)~\cite{kumar2010self}, which can enforce the model dynamically self-adjusting to the actual learning order according to the training loss. 
In this way, our CL-based GET model can make the learning process move towards a better global optimum upon the heterogeneous data to generate high-quality and fine-grained types.
}
Our contributions in this paper are summarized as follows:
\begin{itemize}[noitemsep]
    \item To the best of our knowledge, our work is the first to propose the paradigm of generative entity typing (GET).
    \item We propose to leverage curriculum learning to train our GET model upon heterogeneous data, where the curriculum can be self-adjusted with self-paced learning.
    \item Our extensive experiments on the data of different languages and downstream tasks justify the superiority of our GET model.
\end{itemize}

\section{Related Work}

\paragraph{Classification-based Entity Typing}
\ysy{The traditional classification-based entity typing methods can be categorized into three classes. 
1) \textit{Coarse-grained entity typing} methods~\cite{weischedel2005bbn,tokarchuk2021investigation} assign mentions to a small set of coarse types;
2) \textit{Fine-grained entity typing (FET)} methods~\cite{yuan2018otyper,onoe2021modeling} classify mentions into more diverse and semantically richer ontologies;
3) \textit{Ultra-fine entity typing} methods~\cite{choi2018ultra,ding2021prompt,dai2021ultra} use a large open type vocabulary to predict a set of natural-language phrases as entity types based on texts.
However, FET and ultra-fine entity typing methods hardly perform satisfactorily due to the huge predefined type set. 
They also hardly handle few-shot and zero-shot issues.
Comparatively, our GET model can generate high-quality multi-granularity types even beyond the predefined set for the given entity mentions.}

\paragraph{Concept Acquisition}
Concept acquisition is very related to entity typing which also aims to obtain the types for the given entities, since entity types are often recognized as concepts. 
Concept acquisition can be categorized into the extraction-based or generation-based scheme. 
The extraction scheme cannot acquire concepts not existing in the given text~\cite{yang2020clinical}. 
The existing approaches of concept generation~\cite{zeng2021enhancing} focus on utilizing the existing concept taxonomy or knowledge bases to generate concepts but neglect to utilize the large corpus. Our GET model can also achieve text-based concept generation.

\paragraph{Curriculum Learning}
According to the curriculum learning (CL) paradigm, a model is first trained with the easier subsets or subtasks, and then the training difficulty is gradually increased~\cite{bengio2009curriculum} to improve model performance in difficult target tasks, such as domain adaption~\cite{2021A} and training generalization~\cite{huang2019self}.
The existing CL methods can be divided into predefined CL (PCL)~\cite{bengio2009curriculum} and automatic CL (ACL)~\cite{kumar2010self}. 
PCL divides the training data by the difficulty level with prior knowledge, while ACL, such as self-paced learning (SPL), measures the difficulty according to its losses or other models.

\section{Methodology}\label{sec:Methodology}

\begin{figure}[t]
    \centering
    \includegraphics[width=\columnwidth]{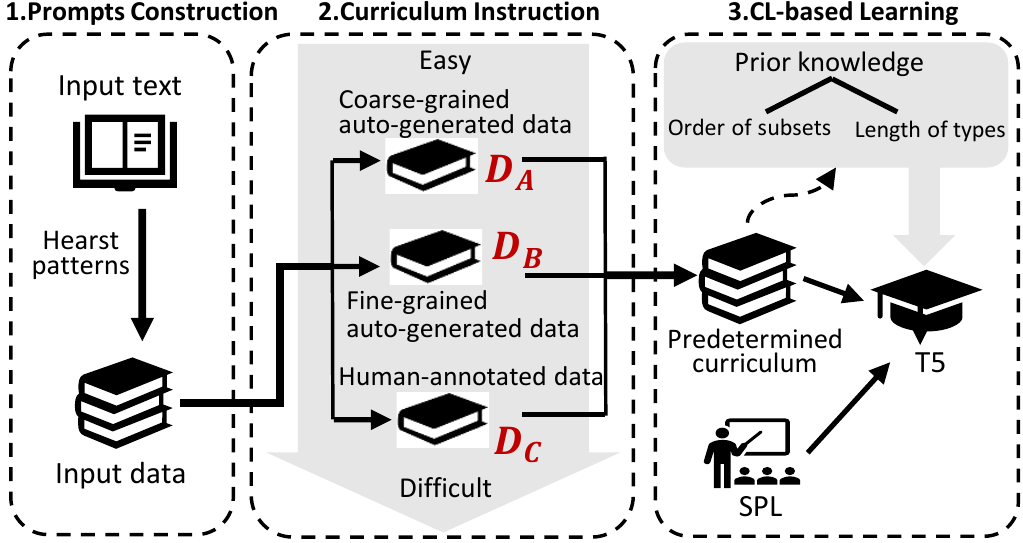}
    \caption{Our PLM-based GET framework trained with curriculum learning.}
    \label{fig:Model}
\end{figure}
In this section, we first formalize our task in this paper and overview the framework of our GET model. Then, we introduce the details of model implementation.

\subsection{Task Formalization}
Given a piece of text $X$ and an entity mention $M$ within it, the task of {\bf generative entity typing (GET)} is to generate multiple types $TS=\{T_1, T_2, ..., T_K\}$, where each $T_k (1\leq k\leq K)$ is a type for $M$ w.r.t. the context of $X$. 

\subsection{Framework}
\label{sec:Framework}
As most of the previous entity typing models~\cite{choi2018ultra,lee2020chinese,gong2021abusive}, our GET model is also trained upon the heterogeneous data consisting of a small portion of human-annotated data and a large portion of auto-generated data, due to the difficulty and high cost of human annotation. 
We will introduce how to obtain our auto-generated data in Section~\ref{sec:autoData}. 
The framework of our model learning includes the following three steps, as shown in Figure~\ref{fig:Model}.
\begin{enumerate}
    \item \textit{Prompt Construction}: To better leverage the knowledge obtained from the pre-training of PLM, we employ the prompt mechanism~\cite{liu2021pre} to guide the learning of our PLM-based GET model;
    \item \textit{Curriculum Instruction}: As a key component of CL, the curriculum instruction is responsible for measuring the difficulty of each sample in the heterogeneous training data, and then designing a suitable curriculum for the model training process;
    \item \textit{CL-based Learning}: In this step, our PLM-based GET model is trained with the designed curriculum, which is capable of adjusting its learning progress dynamically through self-paced learning (SPL).
\end{enumerate}

\subsection{Prompt Construction}
\ysy{To generate the types of given entities by a PLM, we construct the prompts in cloze format from the Hearst patterns listed in Table~\ref{tb:hearst_pattern}.}
Specifically, each input text $X$ including an entity mention $M$ is concatenated with a cloze prompt constructed with $M$, and the PLM is asked to fill the blank within the cloze prompt.
Recall the example in Figure~\ref{fig:Example}, the original text ``\textit{In the early 1980s, P \& G tried to launch here a concentrated detergent under the Ariel brand name that it markets in Europe}'' can be concatenated with a cloze prompt such as ``P \& G is a \underline{\qquad}'' to construct an input prompt for the PLM, which predicts ``large detergent company", ``large detergent manufacturer" and ``company" as the types for \textit{P \& G} to fill the blank.

\begin{table}[!htb]
\centering

\begin{tabular}{|c|c|}
\hline
M is a \underline{\qquad}           & \underline{\qquad} such as M    \\ \hline
M is one of \underline{\qquad}      & \underline{\qquad} especially M \\ \hline
M refers to \underline{\qquad}      & \underline{\qquad}, including M \\ \hline
M is a member of \underline{\qquad} &                                                                 \\ \hline
\end{tabular}
    \caption{Prompts constructed from Hearst patterns.}
    \label{tb:hearst_pattern}
\end{table}

\subsection{Curriculum Instruction}\label{sec:Curriculum Instruction}
Curriculum instruction is the core issue of CL, which requires estimating the difficulty of samples in terms of model learning, to decide the order of using samples for model training.

\begin{figure}[t]
    \centering
    \includegraphics[width=0.95\columnwidth]{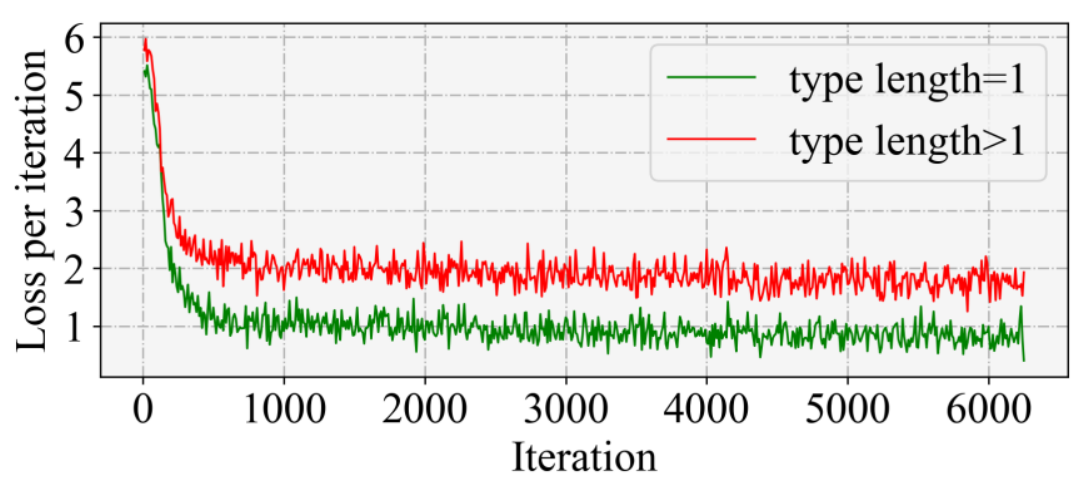} 
    \caption{The landscapes of CE loss comparison for the two training subsets with type length=1 and type length$\ge$2.}
    \label{fig:difficult_measure}
\end{figure}

For our specific PLM-based GET model, we argue that the difficulty of a sample in terms of model learning greatly depends on the granularity of its type, which could be roughly measured by the length of the type term.
To prove this, we generate two subsets of training samples in the same size from the auto-generated data according to their type length: one subset with type length=1, and the other subset with type length$\ge$2\footnote{Here we chose length=1 and length$\ge$2 to divide the training data since most of the types are no longer than 2}.
Then, we train a typical PLM, T5~\cite{raffel2019exploring} for one epoch in the two subsets and record the landscapes of loss. 
As observed in Figure~\ref{fig:difficult_measure}, the length$\ge$2 subset has lower converge and higher cross-entropy (CE) losses than the length=1 subset when training converges, which shows that it is more difficult for the PLM to fit the training samples of fine-grained types.

Based on this observation, we partition the auto-generated data into two subsets, i.e., the subset with one-word types (denoted as $D_A$), and the subset with types of more than one word (denoted as $D_B$), and $D_A$ is used for model training earlier than $D_B$.
The human-annotated data (denoted as $D_C$) is finally used, as it usually contains many ultra fine-grained types annotated by human annotators, which is harder for model learning.
\ysy{
For easier presentation later, we denote the whole training data as $D = D_A\bigcup D_B\bigcup D_C =  \{<X^{(i)},M^{(i)},TS^{(i)}>\}_{i=1}^{N}$, where $N$ is the training sample size and a sample is denoted as $D^{(i)}_k =$  $<X^{(i)},M^{(i)},T^{(i)}_k>$.
Based on the fixed curriculum instruction, CL can control the order in which data are used for model training, i.e., from $D_A$ to $D_B$ to $D_C$.
}

\subsection{CL-based Learning}
\ysy{
\paragraph{T5 Backbone}
To meet the cloze format and generate more fine-grained types, we choose T5~\cite{raffel2019exploring} as the backbone PLM of our GET model.
T5 is an encoder-decoder pre-trained model, which can focus on the entire text and allow multiple words to fill in the blanks. 
To train the T5, in our settings, we define the loss function of sample $D^{(i)}_k$ as,
\begin{equation}\label{eq:sx}
\begin{aligned}
L\big(D^{(i)}_{k}\big) = L_{CE}\big(T_{k}^{(i)},f(X^{(i)},\bm{\theta},M^{(i)})\big)
\end{aligned}
\end{equation}
where $L_{CE}$ is a CE loss function that calculates the cost between the ground truth type $T_{k}^{(i)}$ and the predicted type $f(X^{(i)},\bm{\theta},M^{(i)})$. $\bm{\theta}$ denote the model parameters.

\paragraph{SPL-based Training Process}
Our model training can be performed according to the above predefined curriculum, but the feedback from the learning process is inevitably ignored, which may lead to divergent solutions~\cite{jiang2015self}.
As an alternative, we adopt self-paced learning (SPL) to enforce the model to self-adjust the curriculum according to the feedback from the training loss.
Formally, we define the objective of our CL as, 
\begin{equation}\label{eq:s2}
\begin{aligned}
\min \limits_{\bm{\theta},\bm{v}} E(\bm{\theta},\bm{v};\lambda) &= \sum_{i=1}^N \sum_{k=1}^{K^{(i)}} v^{(i)}_{k}L\big(D^{(i)}_{k}\big)+g(\bm{v};\lambda)
\end{aligned}
\end{equation}
where the binary variable $v_k^{(i)}\in [0,1]$ indicates whether sample $D_k^{(i)}$ should be incorporated into the calculation of the objective. Specifically,
\begin{equation}\label{eq:s3}
 v^{(i)}_{k}=\left\{
\begin{aligned}
1  & &  {L\big(D^{(i)}_{k}\big) < \lambda} \\
0  & &  {L\big(D^{(i)}_{k}\big) \geq \lambda}
\end{aligned}
\right.
\end{equation}
and
\begin{equation}\label{eq:s4}
 g(\bm{v};\lambda) = -\lambda\sum_{i=1}^N\sum_{k=1}^{K^{(i)}} v^{(i)}_{k},
\end{equation}
where $K^{(i)}$ is the number of types for $<X^{(i)},M^{(i)},TS^{(i)}>$, and $\lambda$ is the ``age" of SPL to control the learning pace. 
The regularizer $g(;)$ is the binary self-paced function used to avoid over-fitting~\cite{jiang2014easy}. 

In the training process, ``easy" samples with small losses are used first for training. 
We update $\lambda = \mu\lambda$ to increase $\lambda$ gradually, where $\mu > 1$ is the step size. 
With the growth of $\lambda$, more samples with larger losses are gradually incorporated into model training to obtain a more ``mature" model.
}

\paragraph{Prior Knowledge to Optimize SPL}
As mentioned in Section~\ref{sec:Curriculum Instruction}, we expect that the model is trained orderly by the three subsets according to the predetermined curriculum and generates more fine-grained types. 
However, the order of using the data for model training totally depends on the loss function (Eq. \ref{eq:sx}) in SPL. Thus, SPL is limited in incorporating predetermined curriculum and the type granularity into learning.
Therefore, we treat the predetermined curriculum and the type granularity as prior knowledge to optimize SPL.
Specifically, we increase the weight of the samples with fine-grained types to let the model pay more attention to these data and assign each subset with different weights to ensure that the training process is executed according to the predetermined curriculum. 
In particular, given the sample $D^{(i)}_k$, we define its weight as
\begin{equation}\label{eq:s6}
w(D^{(i)}_{k}) = length(T_{k}^{(i)})*\gamma(D^{(i)}_{k})
\end{equation}
where 
\begin{equation}\label{eq:s5}
 \gamma(D^{(i)}_{k})=\left\{
\begin{aligned}
1  & & \text{if } D^{(i)}_k\in D_A, \\
2  & & \text{if } D^{(i)}_k\in D_B,\\
3  & & \text{if } D^{(i)}_k\in D_C.
\end{aligned}
\right.
\end{equation}
\ysy{
Then, the loss function $L$ in Eq.~\ref{eq:sx} is updated as
\begin{align}\label{eq:s7}
	L_{CE}\big(T_{k}^{(i)},f(X^{(i)},\bm{\theta},M^{(i)})) * w\big(D^{(i)}_{k}\big),
\end{align}
which indicates that a sample with a large weight \dqyang{(more difficult)} can be incorporated later into the training process since its $v_k^{(i)}$ is more likely to be 0 according to Eq.~\ref{eq:s3}. 
We adopt an alternative convex search (ACS) to realize SPL, of which the algorithm is shown in Appendix~\ref{sec:cl_algorithm}. 
We use Adam~\cite{DBLP:journals/corr/KingmaB14} to update the model parameters.
}
\subsection{Type Generation}\label{sec:Type_generation}
When the PLM in our GET model has been trained through the aforementioned CL-based learning process, we use it to generate the types for the given entity mention.
Specifically, we let the PLM fill in the blank of the input text. 
To obtain more diverse candidate types, we apply the beam search ~\cite{reddy1977speech} with beam size as $b$, and select the $b$ most probable candidates. 
Then, we reserve the types with confidence scores bigger than 0.5.

\section{Experiments}
\label{sec:Experiment}
\begin{table*}[t]
 \centering
\begin{tabular}{|c|c|c|c|c|}
\hline

\hline
Dataset             & Type                          & Language & Size of D3 & Size of test set \\ \hline

\hline
BNN~\cite{weischedel2005bbn}                 & Coarse-grained                & English  & 10,000     & 500              \\ \hline
FIGER~\cite{shimaoka2016neural}               & Fine-grained                  & English  & 10,000     & 278              \\ \hline
Ultra-Fine~\cite{choi2018ultra}          & Ultra fine-grained                    & English  & 5,500       & 500              \\ \hline
\multirow{2}{*}{GT~\cite{lee2020chinese}} & \multirow{2}{*}{Multilingual} & English  & 4,750      & 250              \\ \cline{3-5} 
                    &                               & Chinese  & 4,750      & 250              \\ \hline

\hline
\end{tabular}
    \caption{The statistic of different entity typing datasets.} \label{tb:Statistic}
\end{table*}

\begin{table*}[t]
 \centering
\begin{tabular}{|c|cccc|cccc|}
\hline

\hline
\multirow{2}{*}{\textbf{Model}}              & \multicolumn{4}{c|}{\textbf{BNN}}                                                                                                      & \multicolumn{4}{c|}{\textbf{FIGER}}                                                                                                    \\ \cline{2-9} 
                                             & \multicolumn{1}{c|}{\textbf{CT \#}} & \multicolumn{1}{c|}{\textbf{Prec.}}   & \multicolumn{1}{c|}{\textbf{R-Recall}}  & \textbf{R-F1}      & \multicolumn{1}{c|}{\textbf{CT \#}} & \multicolumn{1}{c|}{\textbf{Prec.}}   & \multicolumn{1}{c|}{\textbf{R-Recall}}  & \textbf{R-F1}      \\ \hline

\hline
\citet{zhang2018fine}        & \multicolumn{1}{c|}{555}            & \multicolumn{1}{c|}{58.10\%}          & \multicolumn{1}{c|}{50.49\%}          & 54.03\%          & \multicolumn{1}{c|}{348}            & \multicolumn{1}{c|}{62.00\%}          & \multicolumn{1}{c|}{49.85\%}          & 55.26\%          \\ \hline
\citet{lin2019attentive}     & \multicolumn{1}{c|}{534}            & \multicolumn{1}{c|}{55.90\%}          & \multicolumn{1}{c|}{48.58\%}          & 51.98\%          & \multicolumn{1}{c|}{353}            & \multicolumn{1}{c|}{62.90\%}          & \multicolumn{1}{c|}{50.57\%}          & 56.07\%          \\ \hline
\citet{xiong2019imposing}    & \multicolumn{1}{c|}{558}            & \multicolumn{1}{c|}{58.40\%}          & \multicolumn{1}{c|}{50.75\%}          & 54.31\%          & \multicolumn{1}{c|}{350}            & \multicolumn{1}{c|}{62.30\%}          & \multicolumn{1}{c|}{50.09\%}          & 55.53\%          \\ \hline
\citet{ali2020fine}          & \multicolumn{1}{c|}{697}            & \multicolumn{1}{c|}{73.00\%}          & \multicolumn{1}{c|}{63.43\%}          & 67.88\%          & \multicolumn{1}{c|}{399}            & \multicolumn{1}{c|}{\textbf{71.00\%}} & \multicolumn{1}{c|}{57.08\%}          & 63.29\%          \\ \hline
\citet{chen2020hierarchical} & \multicolumn{1}{c|}{718}            & \multicolumn{1}{c|}{75.20\%}          & \multicolumn{1}{c|}{65.35\%}          & 69.93\%          & \multicolumn{1}{c|}{388}            & \multicolumn{1}{c|}{69.10\%}          & \multicolumn{1}{c|}{55.56\%}          & 61.59\%          \\ \hline
\citet{zhang2021learning}    & \multicolumn{1}{c|}{732}            & \multicolumn{1}{c|}{76.70\%}          & \multicolumn{1}{c|}{66.65\%}          & 71.32\%          & \multicolumn{1}{c|}{394}            & \multicolumn{1}{c|}{70.10\%}          & \multicolumn{1}{c|}{56.36\%}          & 62.48\%          \\ \hline
\citet{lienhancing}          & \multicolumn{1}{c|}{668}            & \multicolumn{1}{c|}{69.90\%}          & \multicolumn{1}{c|}{60.74\%}          & 65.00\%          & \multicolumn{1}{c|}{397}            & \multicolumn{1}{c|}{70.60\%}          & \multicolumn{1}{c|}{56.76\%}          & 62.93\%          \\ \hline
Ours                                         & \multicolumn{1}{c|}{\textbf{875}}   & \multicolumn{1}{c|}{\textbf{82.30\%}} & \multicolumn{1}{c|}{\textbf{79.62\%}} & \textbf{80.94\%} & \multicolumn{1}{c|}{\textbf{444}}   & \multicolumn{1}{c|}{66.20\%}          & \multicolumn{1}{c|}{\textbf{63.52\%}} & \textbf{64.83\%} \\ \hline

\hline
\end{tabular}
\caption{Comparison results of different approaches on the sample test set in coarse-grained and fine-grained entity typing dataset.} \label{tb:traditional_entity_types1}
\end{table*}
\begin{table*}[t]
 \centering
\begin{tabular}{|c|cccc|}
\hline

\hline
\multirow{2}{*}{\textbf{Model}}              & \multicolumn{4}{c|}{\textbf{Ultra-Fine}}                                                                                               \\ \cline{2-5} 
                                             & \multicolumn{1}{c|}{\textbf{CT \#}} & \multicolumn{1}{c|}{\textbf{Prec.}}   & \multicolumn{1}{c|}{\textbf{R-Recall}}  & \textbf{R-F1}      \\ \hline

\hline
\citet{xiong2019imposing}    & \multicolumn{1}{c|}{782}            & \multicolumn{1}{c|}{50.30\%}          & \multicolumn{1}{c|}{24.28\%}          & 32.75\%          \\ \hline
\citet{onoe2019learning} ELMo & \multicolumn{1}{c|}{884}            & \multicolumn{1}{c|}{51.50\%}          & \multicolumn{1}{c|}{27.44\%}          & 35.81\%          \\ \hline
\citet{onoe2019learning} BERT & \multicolumn{1}{c|}{884}            & \multicolumn{1}{c|}{51.60\%}          & \multicolumn{1}{c|}{27.44\%}          & 35.83\%          \\ \hline
\citet{F2020A}               & \multicolumn{1}{c|}{915}            & \multicolumn{1}{c|}{43.40\%}          & \multicolumn{1}{c|}{28.41\%}          & 34.34\%          \\ \hline
\citet{onoe2021modeling}     & \multicolumn{1}{c|}{1039}           & \multicolumn{1}{c|}{52.80\%}          & \multicolumn{1}{c|}{32.26\%}          & 40.05\%          \\ \hline
\citet{liu2021fine}          & \multicolumn{1}{c|}{1042}           & \multicolumn{1}{c|}{54.50\%}          & \multicolumn{1}{c|}{32.35\%}          & 40.60\%          \\ \hline
\citet{dai2021ultra}         & \multicolumn{1}{c|}{1213}           & \multicolumn{1}{c|}{53.60\%}          & \multicolumn{1}{c|}{37.66\%}          & 44.24\%          \\ \hline
Ours                                         & \multicolumn{1}{c|}{\textbf{1275}}  & \multicolumn{1}{c|}{\textbf{87.10\%}} & \multicolumn{1}{c|}{\textbf{39.58\%}} & \textbf{54.43\%} \\ \hline

\hline
\end{tabular}
\caption{Comparison results of different approaches on the sample test set in Ultra-fine entity typing dataset.} \label{tb:traditional_entity_types2}
\end{table*}
\begin{table}[t]
 \centering
\begin{tabular}{|c|ccc|}
\hline

\hline
    \textbf{Dataset}  & \multicolumn{1}{c|}{\textbf{MaNew}} & \multicolumn{1}{c|}{\textbf{MiNew}} & \textbf{R.New}  \\ \hline

\hline
BNN                      & \multicolumn{1}{c|}{4}     & \multicolumn{1}{c|}{100}   & 11.61\% \\ \hline
FIGER                    & \multicolumn{1}{c|}{25}    & \multicolumn{1}{c|}{137}   & 26.81\% \\ \hline
Ultra-Fine               & \multicolumn{1}{c|}{73}    & \multicolumn{1}{c|}{543}   & 42.14\% \\ \hline

\hline
\end{tabular}
    \caption{The number and ratio of new types generated by our model on different datasets.} \label{tb:New}
\end{table}
\begin{table*}[!t]
 \centering
\begin{tabular}{|c|c|cccc|cccc|}
\hline

\hline
\multirow{2}{*}{\textbf{Model}} & \multirow{2}{*}{\textbf{Dataset}}                                                   & \multicolumn{4}{c|}{\textbf{Chinese}}                                                                                                                                         & \multicolumn{4}{c|}{\textbf{English}}                                                                                                                                         \\ \cline{3-10} 
                                &                                                                                     & \multicolumn{1}{c|}{\textbf{CT \#}} & \multicolumn{1}{c|}{\textbf{Prec.}}   &  \multicolumn{1}{c|}{\textbf{R-F1}}      & \textbf{Len.} & \multicolumn{1}{c|}{\textbf{CT \#}} & \multicolumn{1}{c|}{\textbf{Prec.}}   &  \multicolumn{1}{c|}{\textbf{R-F1}}      & \textbf{Len.} \\ \hline

\hline
FT                              & \multirow{4}{*}{\begin{tabular}[c]{@{}c@{}}Auto-\\ generated\\ data\end{tabular}}   & \multicolumn{1}{c|}{690}            & \multicolumn{1}{c|}{84.46\%}          &  \multicolumn{1}{c|}{70.81\%}          & 2.80             & \multicolumn{1}{c|}{870}            & \multicolumn{1}{c|}{75.85\%}          &  \multicolumn{1}{c|}{52.87\%}          & 1.48            \\ \cline{1-1} \cline{3-10} 
PCL                             &                                                                                     & \multicolumn{1}{c|}{646}            & \multicolumn{1}{c|}{91.76\%}          &  \multicolumn{1}{c|}{70.37\%}          & 2.75            & \multicolumn{1}{c|}{864}            & \multicolumn{1}{c|}{85.97\%}          &  \multicolumn{1}{c|}{54.87\%}          & 1.32            \\ \cline{1-1} \cline{3-10} 
SPL w/o PK                      &                                                                                     & \multicolumn{1}{c|}{672}            & \multicolumn{1}{c|}{\textbf{92.18\%}} &  \multicolumn{1}{c|}{72.22\%}          & 2.75            & \multicolumn{1}{c|}{900}            & \multicolumn{1}{c|}{87.12\%}          &  \multicolumn{1}{c|}{56.66\%}          & 1.54            \\ \cline{1-1} \cline{3-10} 
Ours                            &                                                                                     & \multicolumn{1}{c|}{\textbf{714}}   & \multicolumn{1}{c|}{90.04\%}          &  \multicolumn{1}{c|}{\textbf{74.18\%}} & \textbf{2.86}   & \multicolumn{1}{c|}{\textbf{928}}   & \multicolumn{1}{c|}{\textbf{87.14\%}} &  \multicolumn{1}{c|}{\textbf{57.84\%}} & \textbf{1.62}   \\ \hline

\hline
FT                              & \multirow{4}{*}{\begin{tabular}[c]{@{}c@{}}Human-\\ annotated \\ data\end{tabular}} & \multicolumn{1}{c|}{383}            & \multicolumn{1}{c|}{72.54\%}          &  \multicolumn{1}{c|}{53.98\%}          & \textbf{2.65}  & \multicolumn{1}{c|}{352}            & \multicolumn{1}{c|}{84.82\%}          &  \multicolumn{1}{c|}{48.82\%}          & 1.72           \\ \cline{1-1} \cline{3-10} 
PCL                             &                                                                                     & \multicolumn{1}{c|}{370}            & \multicolumn{1}{c|}{77.24\%}          &  \multicolumn{1}{c|}{54.01\%}          & 2.64           & \multicolumn{1}{c|}{\textbf{375}}   & \multicolumn{1}{c|}{88.03\%}          &  \multicolumn{1}{c|}{51.62\%}          & 1.69           \\ \cline{1-1} \cline{3-10} 
SPL w/o PK                      &                                                                                     & \multicolumn{1}{c|}{383}            & \multicolumn{1}{c|}{78.64\%}          &  \multicolumn{1}{c|}{55.59\%}          & 2.61           & \multicolumn{1}{c|}{370}            & \multicolumn{1}{c|}{90.46\%}          &  \multicolumn{1}{c|}{51.53\%}          & 1.74           \\ \cline{1-1} \cline{3-10} 
Ours                            &                                                                                     & \multicolumn{1}{c|}{\textbf{409}}   & \multicolumn{1}{c|}{\textbf{83.64\%}} &  \multicolumn{1}{c|}{\textbf{59.28\%}} & 2.63           & \multicolumn{1}{c|}{373}            & \multicolumn{1}{c|}{\textbf{90.75\%}} &  \multicolumn{1}{c|}{\textbf{51.88\%}} & \textbf{1.82}   \\ \hline

\hline
\end{tabular}
 \caption{Performance comparisons of our model and its variants on the auto-generated and human-annotated test set.} \label{tb:abstract}
\end{table*}

In this section, we verify the advantages of our GET model over the classification-based entity typing models through our experiments. 
We also explore the role of CL in guiding PLM-based type generation with different language data. 
We further display the effectiveness of our generated entity types in two downstream tasks. 

\subsection{Datasets}
As we mentioned before, due to the expensive manual labeling of fine-grained types, the training dataset consists of a large portion of auto-generated data and a small portion of human-annotated data.
Since PLMs have more difficulty fitting the training samples of fine-grained types (elaborated in Figure~\ref{fig:difficult_measure} in Section~\ref{sec:Curriculum Instruction}), we partition the auto-generated data into two subsets, and denote the subset with one-word types as $D_A$ while the other as $D_B$. 
Furthermore, human-annotated data with ultra fine-grained types is denoted as $D_C$.

\paragraph{Auto-generated Data} \label{sec:autoData} 
The auto-generated data used in our model is obtained from the abstracts of entities on Wikipedia~\cite{vrandevcic2014wikidata}. 
Specifically, we collect the abstract texts and their hyperlinks pointing to the web pages of mentioned entities, from which the type labels of these mentioned entities can be obtained. 
In this way, the obtained type labels are more consistent with the contexts of entities, and thus of much higher quality than those auto-generated with distant supervision~\cite{gong2021abusive}.

To construct $D_A$ and $D_B$ from the auto-generated data, we collect 100,000 Chinese and English abstracts from Wikipedia, from which we randomly select 500 samples as our test set, and the rest are used as the training set. 
Then we split the training set into two subsets $D_A$ and $D_B$, according to the length of the types as mentioned in Section~\ref{sec:Curriculum Instruction}.

\paragraph{Human-annotated Data}
\ysy{
To demonstrate that GET is superior to the classification-based approaches, we collect the human-annotated data from four different entity type datasets.
The statistics of them are listed in Table~\ref{tb:Statistic}.
We compare our model with baselines on BNN, FIGER and Ultra-fine to demonstrate the superior performance of GET on entity typing.
GT dataset is used to evaluate the effectiveness of CL upon the texts of different languages and heterogeneous data.
Please note that we do sample the test set to reduce the cost of manual evaluation on assessing whether the newly-generated types are correct. 
However, the results from the baselines and our model are evaluated in the same test instances.
}

\subsection{Evaluation Metrics}
The detailed information about the baselines and some experiment settings is shown in Appendix~\ref{sec:Baselines}. 
Please note that the baselines on BNN and FIGER are different from those on Ultra-Fine.

For BNN and FIGER, we only reserve the type in the predefined type set with the highest probability predicted by the model since there is only one golden label in the dataset. 
Ultra-fine entity typing aims to predict a set of natural-language phrases that describe the types of entity mentions based on texts. 
Therefore, we reserve the types with the probability bigger than 0.5 for all models. 
Please note that previous work adopts a classification paradigm, while our GET model can generate new types not existing in the predefined type set. 
Therefore, annotators are invited to assess whether the generated types are correct. 
The annotation detail is shown in Appendix~\ref{sec:Human_Assessment}.

\ysy{We record the number of correct types assigned to the entity mentions (CT \#) and strict macro-averaged precision (Prec.).
Obviously, it is impossible to know all correct types generated based on the input text in advance due to incalculable search space. 
Therefore, we report the relative recall.
Specifically, suppose CTs \# is the total number of new types obtained by all models. 
Then, the relative recall (R-Recall) is calculated as CT \# divided by  CTs \#. 
Accordingly, the relative F1 (R-F1) can also be calculated with Prec. and R-Recall.}
In addition, we also record the average length of types (Len.) under different training strategies to investigate the effectiveness of CL and prior knowledge.

\subsection{Overall Comparison Results}
The comparison results on traditional entity typing and Ultra-fine entity typing are shown in Table~\ref{tb:traditional_entity_types1} and Table~\ref{tb:traditional_entity_types2}~\footnote{
The statistical significance test and data size analysis are provided in Appendix~\ref{sec:analysis}}.
The tables show that our model (Ours) achieves consistent performance improvements on these datasets. 
For BNN, our model significantly improves Prec. and covers more entity types. 
For FIGER, our model generates more types than other baselines. 
For Ultra-Fine, the existing models based on the classification paradigm are extremely difficult to select the appropriate types from the large predefined type set. 
Comparatively, our GET model has no classification constraint since it transforms multi-classification into a generation paradigm that is more suitable for PLMs.
Therefore, our model greatly improves the precision of Ultra-Fine and covers more types of entities.

We further display the capability of our model to generate new types beyond the predefined type set, as shown in Table~\ref{tb:New}. 
In the table, MaNew (Macro-New) is the total number of generated types beyond the predefined type set, MiNew (Micro-New) is the total number of generated types beyond the human-annotated type set (i.e., golden labeled set $TS$) of each instance, and R.New is the ratio of new generated types per sample. 
The listed results are counted upon the test sets in different datasets, from which we find that our model can generate abundant types that are not in the golden label set, thereby increasing the diversity of entity types.

\begin{figure}[t]
    \centering
    \includegraphics[width=0.9\columnwidth]{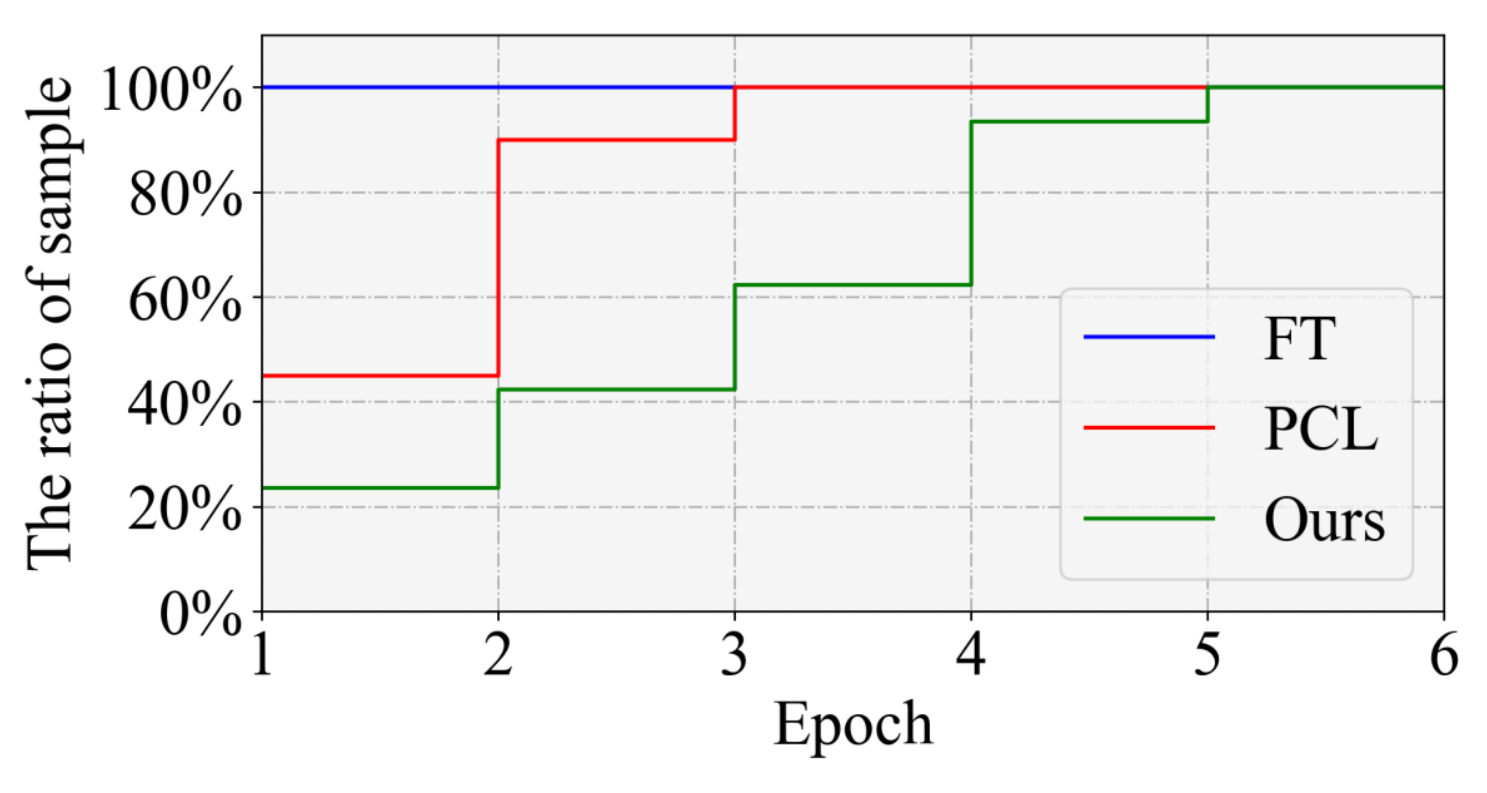}
    \caption{The ratio of training samples of different learning strategies in each epoch.}
    \label{fig:step}
\end{figure}

\begin{figure}[t]
    \centering
    \includegraphics[width=0.9\columnwidth]{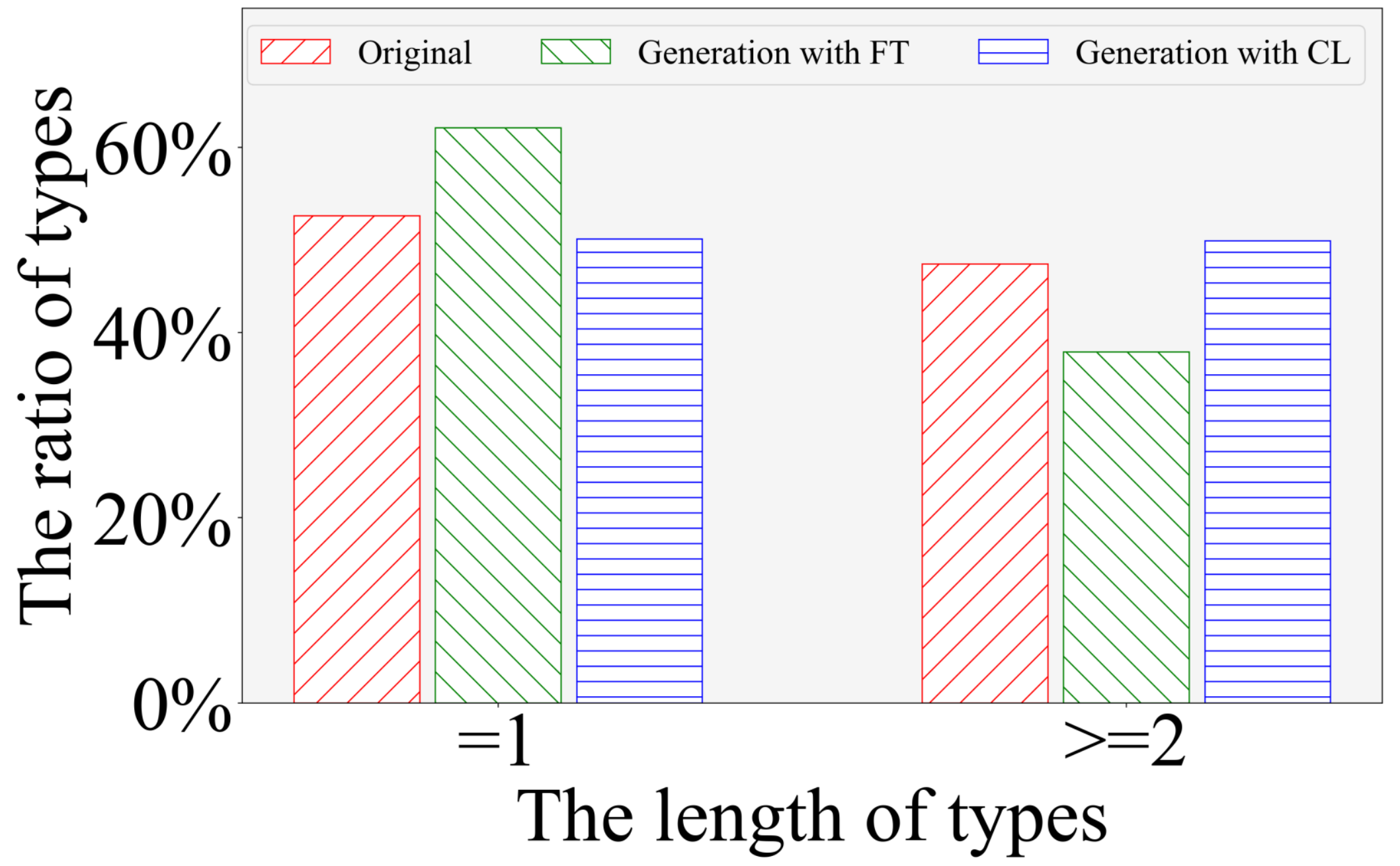} 
    \caption{The ratio of the types of length=1 and length $\ge$ 2 in the original training dataset, and the types generated by our GET model with fine-tuning (FT) or curriculum learning (CL).}
    \label{fig:Distribution}
\end{figure}

\subsection{Effectiveness of CL}
We further compare our model with the following ablated variants to verify the effectiveness of CL. 
\textbf{FT} is fine-tuned directly with training data without CL;
\textbf{PCL} adopts Baby Step~\cite{bengio2009curriculum} instead of SPL, which inputs the subsets into the model in turn according to a fixed curriculum, but ignores the feedback from the learning process;
\textbf{SPL w/o PK} adopts SPL but ignores prior knowledge in training.

In order to demonstrate the performance of the compared models on Chinese and English data, we only consider  $D_C$ of GT in our ablation studies. 
Since the models are designed toward heterogeneous training data, we investigate their performance on both the auto-generated test set and human-annotated test set (in GT), which is displayed in Table~\ref{tb:abstract}. 
Please note that we only report the Prec. and R-F1 due to the limited space.
Based on the results, the superiority of PCL and SPL over FT verifies the advantage of CL over the general training strategy, while the superiority of SPL w/o PK over PCL verifies the effectiveness of SPL. 
Furthermore, our GET model performs well on type generation upon abstract texts (auto-generated data) and common free texts (human-annotated data).

To explore the reason for the advantage of SPL, we record the ratio of incorporated training samples in each epoch. 
As shown in Figure~\ref{fig:step}, SPL gradually incorporates the whole training data to train the model. 
The training on the former subsets can be regarded as a pre-training process that helps model optimization and regularizes the training on the latter subsets. 
Thus, SPL can better guide the model to find a global minimum loss and make it more generalizable.

\subsection{Effectiveness of Prior Knowledge}
\ysy{From the Table~\ref{tb:abstract}, we also find that Ours can generate more fine-grained types than SPL w/o PK.
Without the prior knowledge, SPL only relies on the self-judgment of the model and treats all the selected samples equally, which ignores the data heterogeneity and type granularity during training and harms the model performance.

As shown in Figure~\ref{fig:Distribution}, compared to the original training dataset, there are more coarse-grained types (length=1) than fine-grained types (length$\ge$2) generated by the directly fine-tuned GET model (Generation with FT), while the GET model with CL can generate more fine-grained types of the almost same ratio as the coarse-grained types. 
It is because that the prior knowledge about the type length is considered to re-weight the importance of samples, making the model pay more attention to fine-grained types. 
Thus, more fine-grained and high-quality types are generated. Based on these results, we believe that combining prior knowledge with SPL is an excellent way to optimize CL. 

We also explore the influence of different $\lambda$ and $\mu$ which is shown in Appendix~\ref{sec:Parameter}.}

\begin{table}[t]
 \centering
\begin{tabular}{|l|c|c|c|}
\hline

\hline
\multicolumn{1}{|c|}{\textbf{Method}} & \textbf{Prec.}            & \textbf{Recall}           & \textbf{F1}            \\ \hline

\hline
No type                  & 72.92\%          & 72.70\%          & 72.47\%          \\ \hline
types (KG)               & 73.99\%          & 73.17\%          & 73.30\%          \\ \hline
types (Gen.)             & \textbf{74.51\%} & \textbf{73.41\%} & \textbf{73.53\%} \\ \hline

\hline
\end{tabular}
    \caption{Performance of short text classification based on Bi-LSTM without/with different external knowledge on NLPCC2017 dataset.} \label{tb:NLPCC2017}
\end{table}
\begin{table}[t]
 \centering
\begin{tabular}{|c|l|c|}
\hline

\hline
\textbf{Dataset}                                                                    & \multicolumn{1}{c|}{\textbf{Method}} & \textbf{F1}          \\ \hline

\hline
\multirow{2}{*}{\begin{tabular}[c]{@{}c@{}}AIDA\\ CoNLL-YAGO\end{tabular}} & triples (KG.)                            & 94.58\%          \\ \cline{2-3} 
                                                                           & triples (Gen.)                     & \textbf{94.92\%} \\ \hline
\multirow{2}{*}{ACE 2014}                                                  & triples (KG)                             & 89.74\%          \\ \cline{2-3} 
                                                                           & triples (Gen.)                     & \textbf{90.54\%} \\ \hline

\hline
\end{tabular}
    \caption{Performance of entity linking model DCA-SL with different external knowledge.} \label{tb:entity_linking}
\end{table}

\subsection{Applications}
We further conduct experiments on the task of short text classification and entity linking to prove that the types generated by our model can promote the downstream tasks.

\paragraph{Short Text Classification.}
Existing short text classification approaches ~\cite{chen2019deep} directly use KG as external knowledge to improve model performance. 
However, how to choose the context-consistent types for the roles that the entities play in the text relation is still a problem, which may lead to unsatisfactory results. 
GET can generate context-consistent types for entities, and thus it can be adopted to promote the classification performance. 
We conduct our experiments in the NLPCC2017 dataset~\footnote{\url{http://tcci.ccf.org.cn/conference/2017/taskdata.php}}, the Chinese news title dataset with 18 classes (e.g., entertainment, game, food). 
We first use an NER model to identify entities in the text and directly apply our model upon NLPCC2017 dataset to generate types for the entities. 
Then we choose Bi-LSTM to achieve classification. 
we also collect corresponding types of entities in the representative KG \textit{CN-DBpedia} for comparison.
The results in Table~\ref{tb:NLPCC2017} show that external knowledge enhances the classification performance, and the types generated by our GET model are more effective than those directly obtained from KG.

\paragraph{Entity Linking.}
The representative entity linking model DCA-SL~\cite{yang2019learning} adopts the entity description and triples in KG as an external knowledge to enhance the performance of the model in the entity linking task. 
To prove that the types generated by our model are of high quality, we first adopt our model to generate types for entities based on texts. 
Then we replace the types in the original triples in KGs with the types we generated. 
From Table~\ref{tb:entity_linking} we find that the generated types by our model can improve the entity lining performance of DCA-SL effectively, indicating that the generated types are of high quality and meaningful.

\section{Conclusion}
In this paper, we propose a novel generative paradigm for entity typing, which employs a generative PLM trained with curriculum learning. 
Compared with the traditional classification-based entity typing methods, our generative entity typing (GET) model can generate new types beyond the predefined type set for given entity mentions. 
Our extensive experiments on several benchmark datasets justify that the curriculum learning with SPL and the prior knowledge of type length and subset order help our model generate more high-quality fine-grained types.

\setlength\tabcolsep{4pt}
\begin{table}[t]
 \centering
 \small
\begin{tabular}{|c|c|c|}
\hline

\hline
\textbf{Input Text}                                                                                                                                                        & \textbf{Golden}                                                                                      & \textbf{Generated}                                                                                    \\ \hline

\hline
\begin{tabular}[c]{@{}c@{}}He was capped 42 \\ times and scored 8 \\ goals for \textbf{\underline{Sweden}}, \\ and he played at the \\ 2002 FIFA World Cup\end{tabular}                         & nation                                                                                               & \begin{tabular}[c]{@{}c@{}}nation,\\ nordic country,\\ scandinavian \\ country\end{tabular}           \\ \hline
\begin{tabular}[c]{@{}c@{}}\textbf{\underline{The Audit Bureau of}} \\ \textbf{\underline{Circulations}} \\ was formed in 1914 \\ to verify publication \\ circulation figures and \\ track media rates\end{tabular} & \begin{tabular}[c]{@{}c@{}}administration,\\ organization\end{tabular}                               & \begin{tabular}[c]{@{}c@{}}organization,\\ government \\ agency,\\ investigative \\ service\end{tabular} \\ \hline
\begin{tabular}[c]{@{}c@{}}Chapman retired from \\ playing \textbf{\underline{hockey}} after \\ the 1943 to 1944 \\ hockey season\end{tabular}                                                  & \begin{tabular}[c]{@{}c@{}}tournament,\\ event,contest,\\ game,activity,\\ sport,hockey\end{tabular} & \begin{tabular}[c]{@{}c@{}}sport,\\ contact sport,\\ team sport,\\ winter sport\end{tabular}          \\ \hline

\hline
\end{tabular}
 \caption{Ultra-fine entity typing examples with the corresponding golden labels and generated types. Entity mentions are in bold and underlined.} \label{tb:case_study}
\end{table}
\ysy{\section{Limitations}\label{sec:limitation}
Although we have proven that our GET model can generate high-quality and new types beyond the predefined type set for given entity mentions, it also has some limitations. 
In this section, we analyze these limitations and hopefully advance future work in GET.

\subsection{Uncontrolled Generation}
To delve into the model performance, we compare the types generated by our approach with the golden type labels in Ultra-fine entity typing datasets. 
Table \ref{tb:case_study} lists three examples with the correct types generated by our model and the golden labeled type set of the entities in the Ultra-Fine dataset. 
The first example shows that our model can generate more fine-grained types which may not appear in the golden labeled set. 
The second and third examples demonstrate that although our model can generate new concepts, it may ignore some correct types in the golden label set, e.g., ``administration".
However, enforcing the model by constraint decoding to generate the types in the predefined type set may compromise the flexibility of our model to generate new concepts. 
Therefore, we hope that future work can handle this dilemma with better methods.

\subsection{Type Selection}
As mentioned in Sec.~\ref{sec:Type_generation}, T5 adopts beam search to generate the $b$ most probable types with confidence scores. 
Then we reserve the types with confidence scores larger than the selection threshold. 
However, it can hardly achieve a satisfactory balance to reserve the types by choosing a specific threshold to directly truncate the output of T5. 
If we select a relatively big threshold, we can get more accurate types but may lose some correct types. 
If the recall is preferred, precision might be hurt. 
Therefore, we suggest that future work consider how to achieve a better trade-off between precision and recall.
}

\section*{Acknowledgement}
We would like to thank the anonymous reviewers for their valuable comments and suggestions for this work. This work is supported by the National Natural Science Foundation of China (No.62072323), Shanghai Science, the Science and Technology Commission of Shanghai Municipality Grant (No. 22511105902) and Technology Innovation Action Plan (No.21511100401\& 22511104700). 
\bibliography{anthology}
\bibliographystyle{acl_natbib}

\clearpage
\appendix
\label{sec:appendix}

\section{CL Algorithm}\label{sec:cl_algorithm}
We adopt an alternative convex search (ACS) to realize the SPL of our model training. 
As shown in Algorithm~\ref{alg:algorithm}, we use Adam to update the model parameters.

\begin{algorithm}[!htb]
\caption{Self-paced Learning with prior knowledge for types generation.}
\small
\label{alg:algorithm}
\textbf{Input}: Input training data D, predetermined curriculum $\gamma$, self-paced function $g$, step size $\mu$,  pre-training learning parameters $\bm{\theta}$ , maximum number of iterations $T$ and the number of training epoch $EP$\\
\textbf{Output}: Model parameters $\hat{\bm{\theta}}$
\begin{algorithmic}[1] 
\STATE Initialize $\bm{v}$ and $\lambda$;
\STATE Derive prior knowledge according to $\gamma$ and type length;
\WHILE{$ep < EP$}
\WHILE{$t < T$}
\STATE Compute the objective function $E(\bm{\theta_{t}},\bm{v}; \lambda)$ with prior knowledge;
\STATE Update $\bm{\theta_{t+1}} = Adam(E(\bm{\theta_{t}},\bm{v}; \lambda))$;
\STATE $t = t+1$;
\ENDWHILE
\STATE Record $\hat{\bm{\theta}}$ = $\bm{\theta_{T}}$
\STATE Record $\bm{v}$ = $argmin_{\bm{v}}(E(\hat{\bm{\theta}},\bm{v}; \lambda))$
\STATE $\lambda = \mu\lambda$;
\STATE $ep = ep + 1$;
\ENDWHILE
\STATE \textbf{return} $\hat{\bm{\theta}}$
\end{algorithmic}
\end{algorithm}

\section{Experiment Detail}\label{sec:Baselines}
\subsection{Baselines of Traditional Entity Typing}
Upon traditional entity typing dataset, namely FIGER and BNN, we compare our model with following baselines:
\begin{itemize}
    \item \citet{zhang2018fine}: This approach uses a neural architecture to learn a distributional semantic representation to classify.
    
    \item \citet{lin2019attentive}: This approach proposes a two-step mention-aware attention mechanism to enable the model to focus on the important words in mentions and contexts to improve type classification performance.
    
    \item \citet{xiong2019imposing}: This approach utilizes a graph propagation layer to capture label correlations for type classification.
    
    \item \citet{ali2020fine}: This method adopts edge-weighted attentive graph convolution network to refine the noisy mention representations.
    
    \item \citet{chen2020hierarchical}: Under the undefined case, this approach does not modify the labels in the dataset.
    
    \item \citet{zhang2021learning}: This approach utilizes a probabilistic automatic relabeling method that treats all training samples uniformly to handle noisy samples.
    
    \item \citet{lienhancing}: This approach proposes a novel method based on a two-phase graph network for the Fine-Grained Entity Typing task to enhance the label representations via imposing the relational inductive biases of instance-to-label and label-to-label.

\end{itemize}
\subsection{Baselines of Ultra-Fine Entity Typing}
For Ultra-Fine dataset, we compare our model with the following baselines:
\begin{itemize}

    \item \citet{onoe2019learning}: This approach adopts ELMo and BERT as the encoder to fine-tune on the crowdsourced train split or raw and denoised distantly-labeled data.

    \item \citet{F2020A}: This approach
    proposes a fully hyperbolic model for multi-class multi-label classification, which performs all operations in hyperbolic space.

    \item \citet{onoe2021modeling}: This approach adopts a BERT-based model with box embeddings to capture latent type hierarchies for type classification.
    
    \item \citet{liu2021fine}: This approach discovers and exploits label dependencies knowledge entailed in the data to sequentially reason fine-grained entity labels for type classification.
    
    \item \citet{dai2021ultra}: This approach uses a BERT Masked Language Model to generate weak labels for ultra-fine entity typing to improve the performance of type classification.
    
\end{itemize}
\subsection{Experiment Settings}
Our experiments are conducted on a workstation of dual GeForce GTX 1080 Ti with 32G memory, and the environment of torch 1.7.1. 
We adopted a T5-base with 12 layers and 12 self-attention heads for the English dataset and mT5-small with 8 layers and 6 self-attention heads for the Chinese dataset. 
The hyperparameter settings of training our PLM-based GET are: $\lambda = 0.5$, $\mu = 2$. 
The beam size $b$ is 8. The coefficient weight $\alpha$ in the loss function is 4.

\subsection{Human Assessment}\label{sec:Human_Assessment}
It is impossible to know all newly-generated types apriori. 
Thus human annotators are needed to assess whether the generated types are correct.
We employ two annotators to ensure the quality of the assessment. 
Each predicated type is labeled with 0 or 1 by two annotators, where 0 means a wrong type for the given entity and 1 represents the right type for the given entity. 
If the results from the two annotators are different, the third annotator will be hired for a final check.

\begin{table}[t]
 \centering
 \small

\begin{tabular}{|l|c|c|c|c|}
\hline

\hline
\multicolumn{1}{|c|}{\textbf{Data Size}} & \textbf{CT \#.}     & \textbf{Prec.}        & \textbf{Recall}           & \textbf{F1}            \\ \hline

\hline
50\%         & 1040        & 81.95\%          & 32.29\%          & 46.32\%          \\ \hline
100\%      & \textbf{1275 }     & \textbf{87.10\%}          & \textbf{39.58\% }        & \textbf{54.43\%}          \\

\hline
\end{tabular}
    \caption{Model Performance with different sizes of training data in Ultra-fine entity typing task. 50\% and 100\% are the proportions of auto-generated data used to train the model.} \label{tb:data_size}
\end{table}
\begin{table}[t]
 \centering
 \small
  
\begin{tabular}{|c|c|c|c|c|c|}
\hline

\hline
\textbf{$\mu$} & \textbf{$\lambda$} & \textbf{CT \#} & \textbf{Prec.} & \textbf{Recall} & \textbf{F1} \\ \hline

\hline
2                               & 0.5                                 & \textbf{373}                    & \textbf{90.75\%}                & \textbf{30.52\%}                 & \textbf{45.68\%}             \\ \hline
2                               & 0.1                                 & 330                             & 80.10\%                         & 27.00\%                          & 40.39\%                      \\ \hline
2                               & 1                                   & 359                             & 86.51\%                         & 29.38\%                          & 43.86\%                      \\ \hline
4                               & 0.5                                 & 328                             & 82.21\%                         & 26.84\%                          & 40.47\%                      \\ \hline
4                               & 0.1                                 & 361                             & 82.42\%                         & 29.54\%                          & 43.49\%                      \\ \hline
4                               & 1                                   & 348                             & 81.88\%                         & 28.48\%                          & 42.26\%                      \\ \hline

\hline
\end{tabular}
 \caption{Our model's performance with different hyper-parameter settings.}  \label{tb:new}
\end{table}
\subsection{Result Confidence}\label{sec:analysis}
We also conduct a statistical significance test~\cite{dror-etal-2018-hitchhikers} to show our experiment results are convincing.
Specifically, we run our method on the test set of the Ultra-fine entity typing dataset twice with different random seeds. 
Then we implement a t-test on the two results with a 0.05 significance level.
The result is not significant (p-value: 0.208) and thus we can not reject the null hypothesis (H0: result1-result2=0, where result$_i$=(CT\#, Prec., R-recall, R-F1)). 
Based on the above hypothesis test, we believe that our experiment results are confident and reproducible.

Besides, we do a run with 50\% auto-generated training data for the Ultra-fine entity typing task and the results are shown in Table~\ref{tb:data_size}. 
We find that our method suffers from a slight performance drop but still outperforms the baselines, which shows the effectiveness of auto-generated data.

\subsection{Parameter Tuning Results}\label{sec:Parameter}
We explore the influence of different $\lambda$ and $\mu$ on the performance of our model, as shown in Table~\ref{tb:new}.

\end{document}